\documentclass[letterpaper, 10 pt, conference]{ieeeconf}
\usepackage[a-2b]{pdfx}
\usepackage{graphicx}
\usepackage{amsmath}
\usepackage{amsfonts}
\usepackage{url}

\IEEEoverridecommandlockouts                              
\title{\LARGE \bf
Optimizing Mission Planning for Multi-Debris Rendezvous Using Reinforcement Learning with Refueling and Adaptive Collision Avoidance
}

\author{Agni Bandyopadhyay$^{1}$ and G\"{u}nther Waxenegger-Wilfing$^{2}$
\thanks{$^{1}$Agni Bandyopadhyay is a Doctoral student at Julius-Maximilians-University W\"{u}rzburg, Germany
        {\tt\small agni.bandyopadhyay@uni-wuerzburg.de}}%
\thanks{$^{2}$G\"{u}nther Waxenegger-Wilfing is with the DLR Lampoldshausen, Germany and a Professor at Julius-Maximilians-University W\"{u}rzburg, Germany
        {\tt\small guenther.waxenegger@dlr.de}}%
}

\begin{document}

\maketitle
\thispagestyle{empty}
\pagestyle{empty}

\begin{abstract}
As Earth's orbital environment becomes increasingly crowded with debris, active debris removal (ADR) missions face significant challenges in ensuring safe operations while minimizing the risks of in-orbit collisions. This study presents a novel framework utilizing reinforcement learning (RL) to enhance adaptive collision avoidance in ADR missions, specifically for multi-debris removal using small satellites. Small satellites have been increasingly adopted in recent years for applications ranging from Earth observation to technology demonstration, owing to their flexibility, cost-effectiveness, and maneuverability. These qualities make them particularly well-suited for dynamic and diverse missions such as ADR.

Building on existing research in multi-debris rendezvous, this work integrates refueling strategies, efficient mission planning, and adaptive collision avoidance to optimize mission planning and spacecraft rendezvous operations. The proposed approach employs a masked Proximal Policy Optimization (PPO) algorithm, allowing the RL agent to dynamically adjust spacecraft maneuvers in response to real-time orbital conditions. Key considerations include fuel efficiency, avoidance of active collision zones, and optimization of dynamic orbital parameters. The RL agent is trained to determine the most efficient sequence for rendezvousing with multiple debris targets, optimizing fuel usage and time while incorporating necessary refueling stops.

Simulated ADR scenarios, derived from the Iridium 33 debris dataset, are used to evaluate the performance of the proposed RL framework. These scenarios encompass diverse orbital configurations and debris distributions, showcasing the system’s adaptability and robustness. Results indicate that the RL-based framework not only reduces collision risks but also significantly improves mission efficiency compared to traditional heuristic approaches.

This research provides a practical and scalable solution for planning complex ADR missions targeting multiple debris. While focused on ADR, the RL-based decision-making framework offers broader applicability to other multi-target rendezvous scenarios, demonstrating its potential to improve space mission planning.

\end{abstract}

\section{INTRODUCTION}

The exponential rise in artificial satellites and orbital debris has fundamentally reshaped the dynamics of low Earth orbit (LEO). With over 30,000 cataloged objects and millions of sub-centimeter fragments, the risk of in-orbit collisions is now a critical concern for both public space agencies and commercial satellite operators \cite{liou2006, esa2023, lewis2020}. Catastrophic events such as the Iridium 33–Cosmos 2251 collision have demonstrated the potential for cascading chain reactions, famously known as the Kessler Syndrome \cite{kessler1978, kelso2009}. These self-sustaining collision cascades can significantly impair space access and orbital sustainability for decades to come.

In response, Active Debris Removal (ADR) is gaining traction as a central pillar of orbital traffic management and long-term sustainability \cite{liou2011, wright2022}. The goal of ADR is to physically remove defunct or high-risk objects from orbit to reduce the likelihood of future collisions. However, designing and executing ADR missions is exceptionally complex. The primary challenges lie in sequencing multi-target rendezvous, optimizing fuel consumption, avoiding time-variant collision threats, and managing limited on-board computational resources. Traditional mission planning techniques for debris visitation sequencing, such as brute-force enumeration, become computationally infeasible as mission complexity grows, while greedy heuristics, although computationally efficient, often yield suboptimal solutions \cite{gutin2008, sharma2016}.

Moreover, recent advances in the deployment of mega-constellations such as Starlink and OneWeb have further complicated the orbital environment, introducing high-density conjunction regions and dynamic traffic patterns \cite{lewis2019}. These developments underscore the urgent need for intelligent, adaptive, and scalable mission planning algorithms.

Reinforcement learning (RL) has emerged as a promising paradigm for tackling sequential decision-making under uncertainty. In particular, RL agents can learn effective policies by interacting with a simulated environment and optimizing for long-term cumulative rewards \cite{sutton2018, zhang2020deep}. Applications of RL in aerospace domains have shown strong performance in control, fault recovery, and autonomous planning \cite{cho2022, chen2021}. Moreover, recent studies demonstrate the growing potential of reinforcement learning for spacecraft mission (re)planning and control under resource constraints, highlighting its applicability to autonomous space operations \cite{retagne2024adaptive, tipaldi2022reinforcement}. In the context of ADR, RL offers the ability to train agents that can adapt in real-time to probabilistic events such as emerging collision zones, system malfunctions, or perturbations.

In this paper, we propose a reinforcement learning-based mission planning framework for multi-debris removal missions with fuel constraints, while simulating collision avoidance. Our approach centers on a masked Proximal Policy Optimization (PPO) algorithm that enables the agent to efficiently sequence debris targets, navigate dynamically generated collision risks, and manage fuel and refueling logistics in real time. The core contributions of our work include:

\begin{itemize}
    \item \textbf{Adaptive collision avoidance}, where probabilistically triggered cuboidal risk zones force the RL agent to replan using detour maneuvers that maintain a minimum 5 km clearance;
    \item \textbf{Refueling logic}, which integrates discrete operational checkpoints that become available only after visiting debris, extending mission lifespan while penalizing premature refuels;
    \item \textbf{Fuel-efficient transfers} using Hohmann maneuvers and ellipse-based avoidance arcs;
    \item \textbf{Custom reward shaping} that balances mission efficiency, safety, and full debris coverage.
\end{itemize}

Our simulation environment incorporates realistic orbital dynamics, stochastic collision modeling, and randomized debris scenarios derived from the Iridium 33 dataset. We train the RL agent over 8 million steps, evaluate it on 100 unique test cases, and benchmark it against both greedy and hybrid planning approaches. The results show that our framework outperforms baseline methods in both safety compliance and mission completeness.

The proposed architecture generalizes beyond ADR, with potential applications in multi-target missions such as on-orbit servicing, collaborative inspection, and asteroid sample return campaigns.

\section{BACKGROUND}

Active Debris Removal (ADR) has emerged as a critical strategy for maintaining the long-term sustainability of operations in Earth's increasingly congested orbital environment \cite{liou2011, esa2023}. With the proliferation of satellites, particularly from large-scale constellations, the risk of orbital collisions has sharply increased, necessitating the development of intelligent, responsive planning systems \cite{lewis2019}.

ADR missions designed for multi-debris removal require spacecraft to identify and rendezvous with multiple debris objects, typically under tight operational constraints. These include limited fuel, onboard computational resources, time bounds, and real-time environmental hazards such as collisions and perturbations. The core planning challenge—determining the optimal sequence and trajectory for visiting debris targets—resembles the classical Traveling Salesman Problem (TSP) \cite{gutin2008}, yet ADR adds complexity through orbital dynamics and uncertainty.

Key distinguishing features of the ADR problem include:

\begin{itemize}
    \item \textbf{Orbital mechanics:} Transfers between targets are governed by time-dependent Keplerian dynamics. Lambert solvers are required for inter-debris transitions \cite{vallado2007}. In our work, we focus on Hohmann transfer arcs for fuel-efficient rendezvous.
    \item \textbf{Dynamic transfer costs:} Delta-v requirements vary significantly with orbital phase, inclination differences, and relative geometries.
    \item \textbf{Operational constraints:} Missions must respect hard limits on fuel and time, while also satisfying safety margins from collision zones.
    \item \textbf{Environmental uncertainty:} Conjunction risks and debris motion are inherently probabilistic, requiring replanning capabilities.
\end{itemize}

\begin{figure}[htbp]
    \centering
    \includegraphics[width=0.8\linewidth]{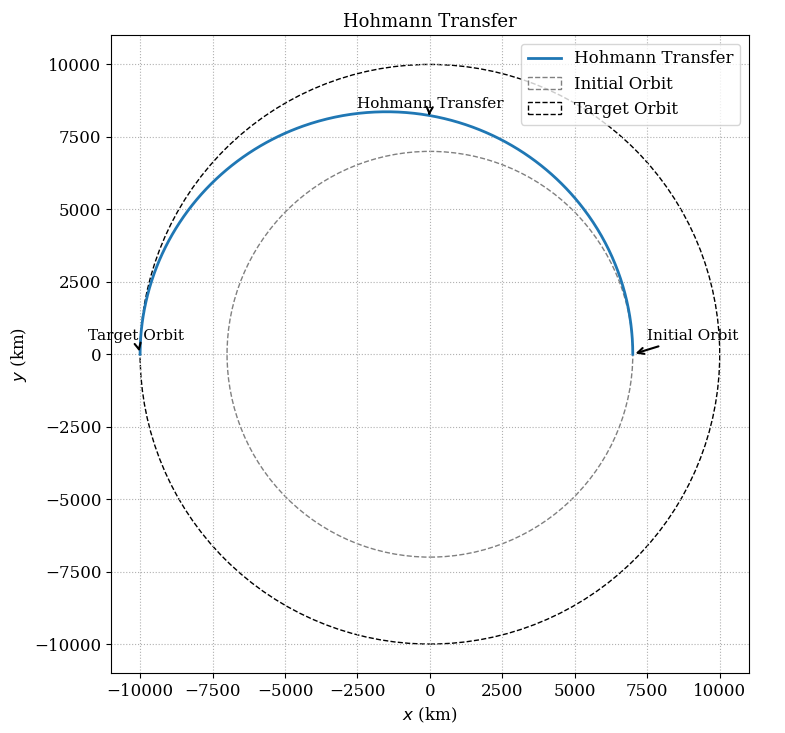}
    \caption{Illustration of a Hohmann transfer maneuver from an initial circular orbit to a higher target orbit. This is the core maneuver model used in our rendezvous planning.}
    \label{fig:hohmann_transfer}
\end{figure}

Traditional approaches to this problem—including greedy heuristics, brute-force search, and genetic algorithms \cite{sharma2016, goldberg1989}—provide only limited adaptability. These methods tend to assume static environments and often cannot handle mid-mission changes such as new collision risks or unexpected fuel depletion.

As the number of active and passive satellites grows, the need for adaptive collision avoidance has become paramount. Fixed safety margins or static detour strategies no longer suffice in dynamic settings like satellite megaconstellations \cite{lewis2019}. In our work, we introduce \emph{cuboidal danger zones} that probabilistically trigger (33\% chance) and necessitate trajectory replanning with a clearance of 5 km.

While refueling theoretically extends mission lifespans, incorporating it into decision policies adds another dimension to the optimization: agents must weigh the cost and benefit of stopping to refuel mid-mission \cite{chen2020orbital}.

Recent studies, including our previous work \cite{bandyopadhayay_2024}, have highlighted the promise of reinforcement learning (RL) as a solution to this complex problem. RL excels in scenarios requiring online sequential decision-making under uncertainty \cite{sutton2018}. By framing ADR as a Markov Decision Process (MDP), we enable the agent to learn not only the optimal debris sequence, but also when to perform refueling and how to respond to collision hazards.

This study expands on that by:
\begin{itemize}
    \item Integrating adaptive collision zones using a probabilistic trigger mechanism;
    \item Embedding refueling checkpoints as learnable decisions;
    \item Employing masked PPO for safety-aware action filtering and policy learning.
\end{itemize}

These elements combine into a unified RL framework designed for generalizable, scalable, and safe ADR planning in dynamic orbital environments.

\subsection{Problem Statement}

The problem of active debris removal (ADR) in congested orbits can be formulated as a sequential decision-making problem under uncertainty, a modified Dynamic TSP \cite{jung2007dynamic}. The primary objective is to maximize the number of debris objects visited by a single satellite within mission constraints, including limited fuel, time bounds, and probabilistic collision risks.

Given a set of $n$ debris objects $\mathcal{D} = \{d_1, d_2, \ldots, d_n\}$, each with known orbital parameters, the satellite must determine an optimal sequence of maneuvers $\mathcal{S} = \{a_1, a_2, \ldots, a_k\}$ where each $a_i$ is one of:
\begin{itemize}
    \item A rendezvous with a debris object $d_j \in \mathcal{D}$, where $j = 1,2, \dotsc, n$,
    \item A refueling action, or
    \item A collision avoidance maneuver (CA Above / CA Below).
\end{itemize}

The trajectory between targets must comply with Keplerian motion and safety constraints. Each maneuver consumes fuel, and the satellite must avoid any collision zones that may dynamically appear with 33\% probability during transfer selection. A mission is deemed successful if the agent:
\begin{itemize}
    \item Maximizes the number of unique debris rendezvoused within the maximum episode duration,
    \item Avoids all collision zones by re-planning using elliptical detours with a minimum 5 km clearance,
    \item Refuels only when eligible (after at least one successful rendezvous),
    \item Stays within the total time and fuel budget.
\end{itemize}

More details can be found in the Appendix.

\section{REINFORCEMENT LEARNING FRAMEWORK}

Reinforcement learning (RL) is well-suited to decision-making problems under uncertainty, especially when the agent must sequentially optimize multiple objectives such as fuel efficiency, target visitation, and collision avoidance \cite{sutton2018}. In this work, we formulate the active debris removal (ADR) problem as a Markov Decision Process (MDP), where the RL agent must learn a policy $\pi(a_t|s_t)$ to select optimal actions $a_t$ given the current environment state $s_t$ at time $t$.

\subsection{State and Action Representation}

The environment state $s_t$ captures the full operational context required for intelligent decision-making in dynamic orbital scenarios. It includes:

\begin{itemize}
    \item Current spacecraft position $\mathbf{r}_t$, velocity $\mathbf{v}_t$, and normalized fuel level $f_t \in [0, 1]$;
    \item Binary visitation mask $\mathbf{m} \in \{0, 1\}^n$ indicating whether each debris $i$ has been visited;
    \item Relative Keplerian elements of each debris object $d_i \in \mathcal{D}$: semi-major axis $a_i$, eccentricity $e_i$, inclination $i_i$, RAAN $\Omega_i$, argument of perigee $\omega_i$, and true anomaly $\nu_i$, for all $i = 1, \dotsc, n$ \cite{vallado2007};

    \item Distance vector $\mathbf{d}_r$ to the nearest refueling stations;
    \item Refueling eligibility flag $r_{\text{eligible}}$ which is set after visiting at least one debris;
    \item Collision risk proximity vector indicating potential danger zones along transfer arcs.
\end{itemize}

\textbf{Action space:} The agent operates in a discrete, masked action space:
\[
\mathcal{A} = \mathcal{A}_\text{debris} \cup \mathcal{A}_\text{refuel} \cup \mathcal{A}_\text{CA}
\]

\begin{itemize}
    \item $\mathcal{A}_\text{debris} = \left\{ i \in [1, n] \mid {m_i = 0 }   \&  \text{ safe}(i) = 1 \right\}$: Valid debris targets not yet visited and currently outside danger zones.
    \item $\mathcal{A}_\text{refuel} = \{ \texttt{Refuel} \}$ if $r_{\text{eligible}} = 1$.
    \item $\mathcal{A}_\text{CA} = \{ \texttt{CA\_Above}(i), \texttt{CA\_Below}(i) \mid i \in [1, n], \text{collision risk on arc to } i \}$: Collision avoidance maneuver selection when a transfer arc intersects a danger zone.
\end{itemize}

Each selected action $a_t$ is interpreted and executed as a rendezvous maneuver, a detour, or a refueling operation. The underlying physics-based propagation (e.g., Hohmann transfers) and safety checks ensure feasibility.

\subsection{Reward Function Design}

The agent is trained to maximize long-term rewards, balancing multiple mission goals. The shaped reward function is defined as:
\begin{equation}
    r_t = \delta_{\text{visit}} -  C_t 
    -T_{penalty}
\end{equation}
where:
\begin{itemize}
    \item $T_{\text{penalty}}$ is the penalty for running out of fuel or total time and is 1;
    \item $\delta_{\text{visit}} = 1$ if a new debris is visited, otherwise 0;
    \item $C_t = 1$ if a collision occurs, otherwise 0;
\end{itemize}

\subsection{Masked PPO Algorithm}

We use Proximal Policy Optimization (PPO) \cite{schulman2017} as the core reinforcement learning algorithm due to its stability and robustness in continuous control domains. PPO optimizes a clipped surrogate objective, encouraging small policy updates to prevent performance collapse. The goal is to minimize the clipped loss function:

\begin{equation}
    L^{\text{CLIP}}(\theta) = \mathbb{E}_t \left[ \min \left( r_t(\theta) \hat{A}_t, \, \text{clip}(r_t(\theta), 1 - \epsilon, 1 + \epsilon) \hat{A}_t \right) \right]
\end{equation}

where:
\begin{itemize}
    \item $\theta$ are the parameters of the policy network,
    \item $\hat{A}_t$ is the estimated advantage at timestep $t$,
    \item $\epsilon$ is the hyperparameter controlling the clip range (typically $0.1$ to $0.2$),
    \item $r_t(\theta) = \frac{\pi_\theta(a_t|s_t)}{\pi_{\theta_{\text{old}}}(a_t|s_t)}$ is the probability ratio between the new and old policies.
\end{itemize}

The expectation $\mathbb{E}_t[\cdot]$ is taken over timesteps sampled from the environment during policy rollouts.  
The clipping function ensures that updates to the policy do not deviate too much from the previous policy, thus stabilizing training.

To further enhance learning stability and ensure safe decision-making, we apply \textit{invalid action masking} \cite{huang2022}. This technique modifies the policy distribution by setting the logits of invalid actions to $-\infty$ prior to applying the softmax:

\begin{equation}
    \pi(a_t|s_t) \propto \text{softmax}(\mathbf{z}_t + \log(\mathbf{v}_t))
\end{equation}

where:
\begin{itemize}
    \item $\mathbf{z}_t$ are the raw output logits of the policy network,
    \item $\mathbf{v}_t$ is a binary mask vector indicating valid (1) and invalid (0) actions.
\end{itemize}

Invalid action masking dynamically restricts the action space at each timestep based on the current mission state, preventing infeasible actions (e.g., visiting already visited debris, refueling without eligibility) from being selected during training and inference.

The agent is trained using action masking to ensure invalid moves (e.g., selecting visited debris, trying to refuel when not allowed, or attempting CA when no collision exists) are excluded from the policy distribution. This results in a dynamic action space at each timestep, tailored to the current mission state.

We train over 10 million steps using Stable-Baselines3 \cite{raffin2021} with distributed sampling and entropy regularization.

This architecture allows the agent to learn both high-level planning (e.g., target sequencing, refueling strategy) and low-level behaviors (e.g., avoidance maneuvers), making it robust across debris densities and mission conditions.

\section{METHODOLOGY AND EVALUATION STRATEGY}

To evaluate the robustness, scalability, and effectiveness of our reinforcement learning (RL) framework, we designed a comprehensive set of simulations and comparative baselines. This section describes our training methodology, evaluation modes, collision avoidance maneuvers, and the simulation stack used to generate realistic debris scenarios.

\subsection{Training Methodology}

The RL agent was trained for over 10 million steps using a masked Proximal Policy Optimization (PPO) algorithm. Training was done with randomized debris configurations for each episode. Each episode simulates a full debris removal mission that begins from a fixed parking orbit and terminates when either:
\begin{itemize}
    \item All debris have been visited;
    \item Fuel is exhausted;
    \item A collision occurs;
    \item The agent runs out of time.
\end{itemize}

Collision risks are dynamically triggered with 33\% probability upon selecting a debris target. When activated, a cuboidal danger zone of size $5 \times 5 \times 5$ km is generated around the collision object. If the planned trajectory intersects this zone, an avoidance maneuver is executed, requiring a detour with a minimum 5 km clearance.

Refueling is permitted only after the agent has completed at least one rendezvous. Once refueled, the eligibility is reset. This models real-world orbital refueling constraints.

\subsection{Collision Avoidance Maneuvers}

When a planned Hohmann transfer intersects a probabilistically triggered collision zone, the agent replans the trajectory using one of two elliptical detour paths:

\begin{itemize}
    \item \textbf{CA\_Above:} The agent shifts to a slightly higher elliptical transfer orbit.
    \item \textbf{CA\_Below:} The agent selects a lower elliptical orbit, reducing proximity to the hazardous zone.
\end{itemize}

\begin{figure}[htbp]
    \centering
    \includegraphics[width=0.47\textwidth]{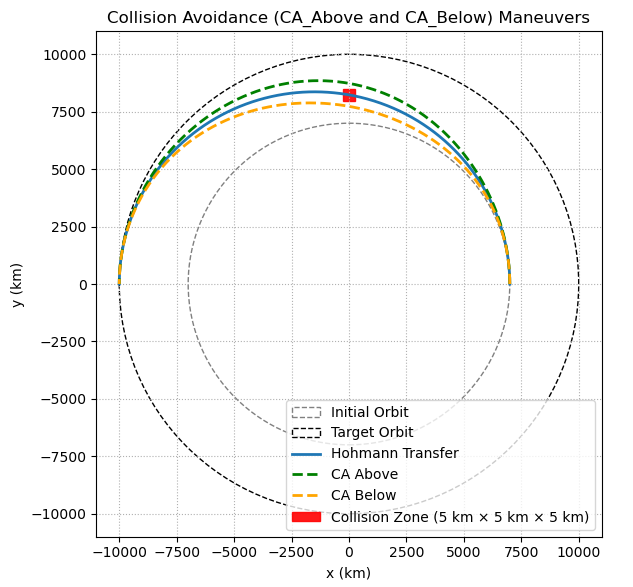}
    \caption{Collision Avoidance Maneuvers: The agent performs CA\_Above or CA\_Below detours around a $5 \times 5 \times 5$ km collision zone to maintain safety margins.}
    \label{fig:collision_avoidance}
\end{figure}

These maneuvers are visualized in Fig.~\ref{fig:collision_avoidance}, which shows the deviation paths relative to the nominal Hohmann arc.

Let $r_1$ denote the radius of the initial orbit and $r_2$ the radius of the target orbit. A standard Hohmann transfer computes an energy-efficient trajectory between two coplanar circular orbits under two-body dynamics assumptions. The semi-major axis $a_H$ of the transfer ellipse and the velocity at the target orbit $v_{t,H}$ are given by:

\[
a_H = \frac{r_1 + r_2}{2}, \quad v_{t,H} = \sqrt{\mu \left( \frac{2}{r_2} - \frac{1}{a_H} \right)}
\]

where $\mu$ is the standard gravitational parameter of Earth.

During collision avoidance events, we modify the standard transfer by introducing a small radial displacement $\Delta r$ to the target orbit. For a CA\_Above maneuver, $r_2$ is increased by $\Delta r$, yielding:

\[
a_{CA} = \frac{r_1 + (r_2 + \Delta r)}{2}, \quad v_{t,CA} = \sqrt{\mu \left( \frac{2}{r_2 + \Delta r} - \frac{1}{a_{CA}} \right)}
\]

Similarly, for CA\_Below, the transfer reduces the target radius by $\Delta r$.

These adjusted trajectories are elliptical detours computed using patched-conic approximations, designed to safely bypass the collision zones by maintaining a minimum clearance of 5 km from the hazardous object. The modified arcs are applied only temporarily during collision events, after which the spacecraft returns to nominal mission planning.

\subsection{Evaluation Modes}

We conduct evaluation across four distinct planning modes to benchmark the RL agent’s behavior:
\begin{enumerate}
    \item \textbf{RL-RL Mode:} The RL agent selects both the visitation sequence and performs collision avoidance, as shown in Figure \ref{fig:rl_rl_mode}.
    \begin{figure}[htbp]
    \centering
    \includegraphics[width=0.45\textwidth]{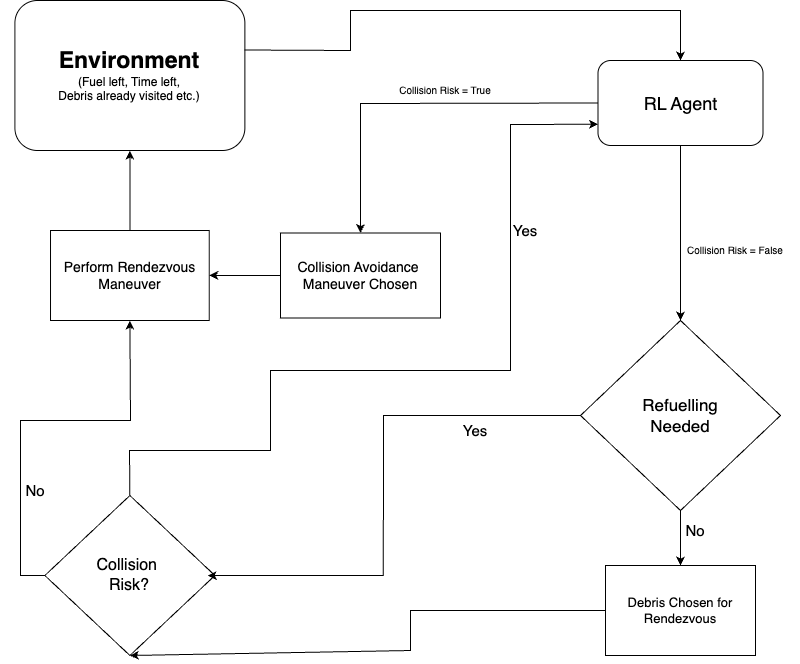}
    \caption{Decision flow in RL-RL mode: The RL agent selects both the debris sequence and handles collision avoidance.}
    \label{fig:rl_rl_mode}
    \end{figure}

    \item \textbf{RL-Greedy Mode:} The RL agent selects the visitation sequence, while collision avoidance is handled using a deterministic greedy planner favoring minimum time, as shown in Figure \ref{fig:rl_greedy_mode}.
    \begin{figure}[htbp]
    \centering
    \includegraphics[width=0.45\textwidth]{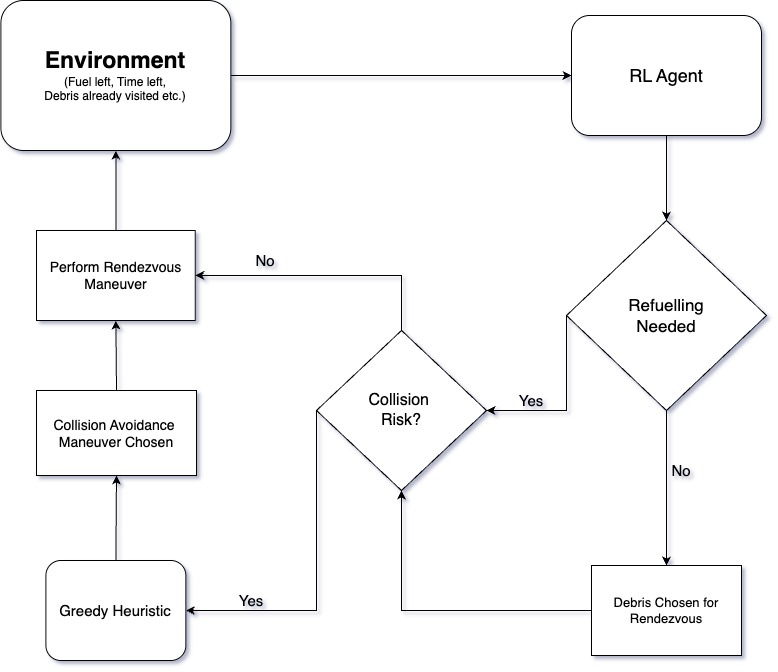}
    \caption{Decision flow in RL-Greedy mode: The RL agent selects the sequence, while a heuristic handles collision avoidance favouring time.}
    \label{fig:rl_greedy_mode}
    \end{figure}

    \item \textbf{Greedy-RL Mode:} A greedy heuristic selects the visitation sequence based on minimum $\Delta v$, while the RL agent is used solely for collision avoidance, as shown in Figure \ref{fig:greedy_rl_mode}.
    \begin{figure}[htbp]
    \centering
    \includegraphics[width=0.45\textwidth]{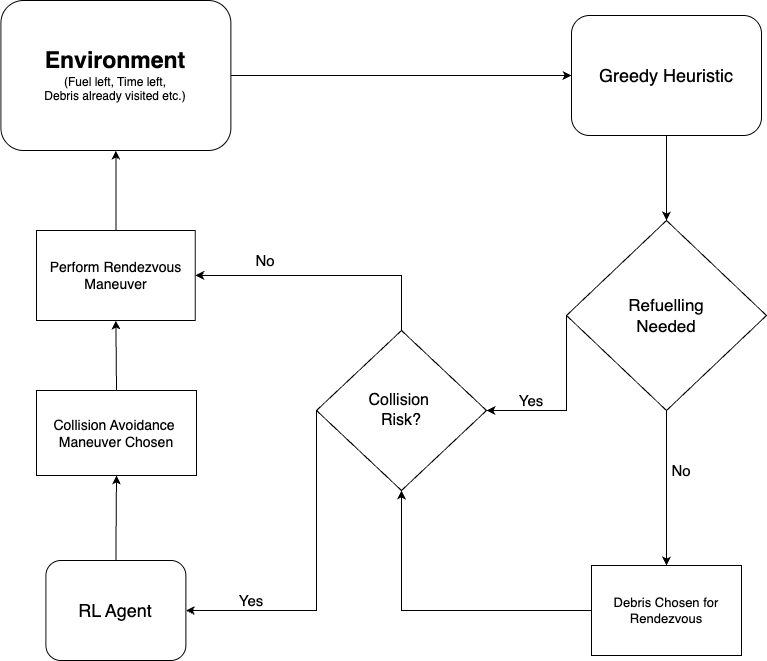}
    \caption{Decision flow in Greedy-RL mode: A heuristic selects the sequence, while the RL agent manages avoidance.}
    \label{fig:greedy_rl_mode}
    \end{figure}

    \item \textbf{Greedy-Greedy Mode:} A fully heuristic approach where the visitation sequence is selected based on minimum $\Delta v$, and collision avoidance is also handled deterministically by a time-minimizing greedy planner, as shown in Figure \ref{fig:greedy_greedy_mode}.
    \begin{figure}[htbp]
    \centering
    \includegraphics[width=0.45\textwidth]{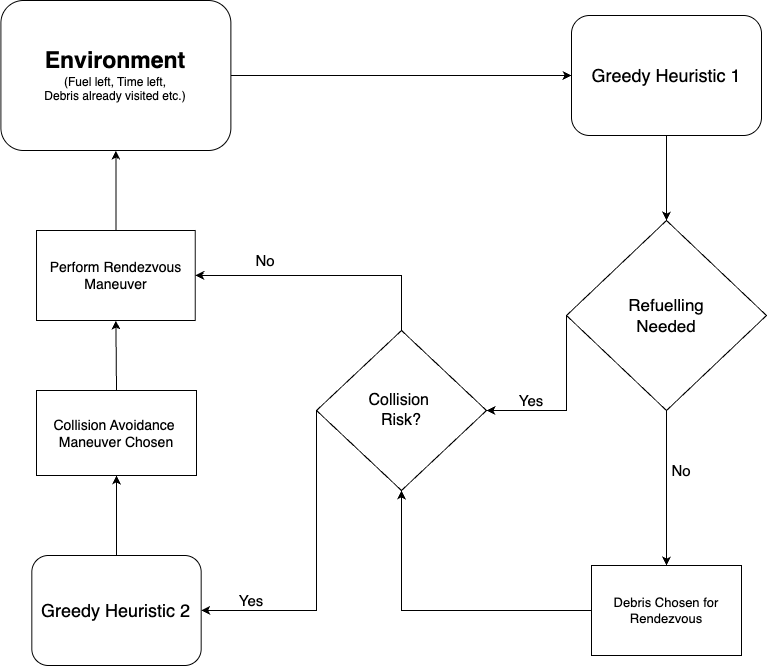}
    \caption{Decision flow in Greedy-Greedy mode: Both visitation sequence and collision avoidance are handled by two separate greedy heuristics.}
    \label{fig:greedy_greedy_mode}
    \end{figure}
\end{enumerate}

\subsection{Metrics and Logging}

For each scenario, we log the following metrics:
\begin{itemize}
    \item \textbf{Average mission time (AMT):} Total time to complete all rendezvous.
    \item \textbf{Debris visited:} Final count of successfully visited debris.
    \item \textbf{Collision rate:} Number of missions ending in collision.
    \item \textbf{Refueling events:} Frequency and timing of refuels.
    \item \textbf{Replanning frequency:} How often the agent triggered avoidance maneuvers.
\end{itemize}

All results are stored in CSV format and used to generate performance visualizations in the Results section.

\subsection{Simulation Tools}

We use the following stack for implementation:
\begin{itemize}
    \item \textbf{Astrodynamics:} Hohmann transfers and orbital propagations are handled using Poliastro and Astropy \cite{poliastro,astropy};
    \item \textbf{Reinforcement Learning:} PPO is implemented via Stable-Baselines3;
    \item \textbf{Collision Detection:} Zone intersections are computed using 3D ellipsoid-cuboid overlap tests;
    \item \textbf{Trajectory Replanning:} Emergency maneuvers follow a heuristic elliptical detour ensuring 5 km buffer.
\end{itemize}

The evaluation runs are executed on an Apple MacBook Pro with M1 Max and 64 GB RAM.

\section{RESULTS AND DISCUSSION}

This section presents a comprehensive analysis of the training process, evaluation outcomes, and emergent behaviors of the proposed RL-based mission planning framework. The results are benchmarked against hybrid and heuristic strategies across multiple randomized debris fields.

\subsection{Training Performance Over Time}

\begin{figure}[htbp]
    \centering
    \includegraphics[width=\linewidth]{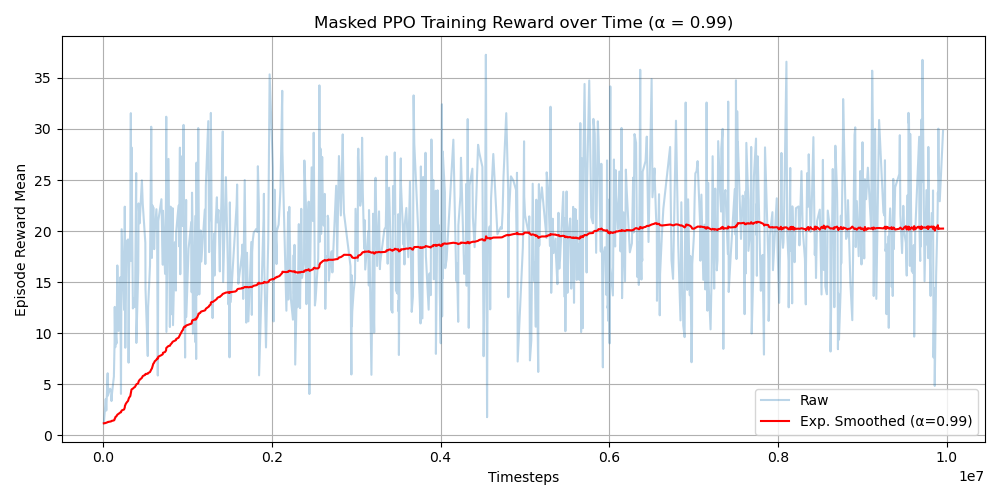}
    \caption{Masked PPO Training Reward over Time ($\alpha = 0.99$). Reward trends upward and stabilizes, indicating learning convergence.}
    \label{fig:ppo_reward}
\end{figure}

Figure~\ref{fig:ppo_reward} shows the agent's reward signal progression over 10 million timesteps. The red line represents the exponentially smoothed curve ($\alpha = 0.99$).

The curve reveals two important trends:
\begin{enumerate}
    \item A steep initial reward gain within the first 1--2 million steps.
    \item A plateauing phase where the agent fine-tunes its mission planning strategies.
\end{enumerate}

The stability of the reward curve post 8 million steps suggests that the policy has converged on a good strategy.

\subsection{Debris Visited Comparison}

\begin{figure*}[t]
    \centering
    \includegraphics[width=0.85\linewidth, height=0.45\textheight]{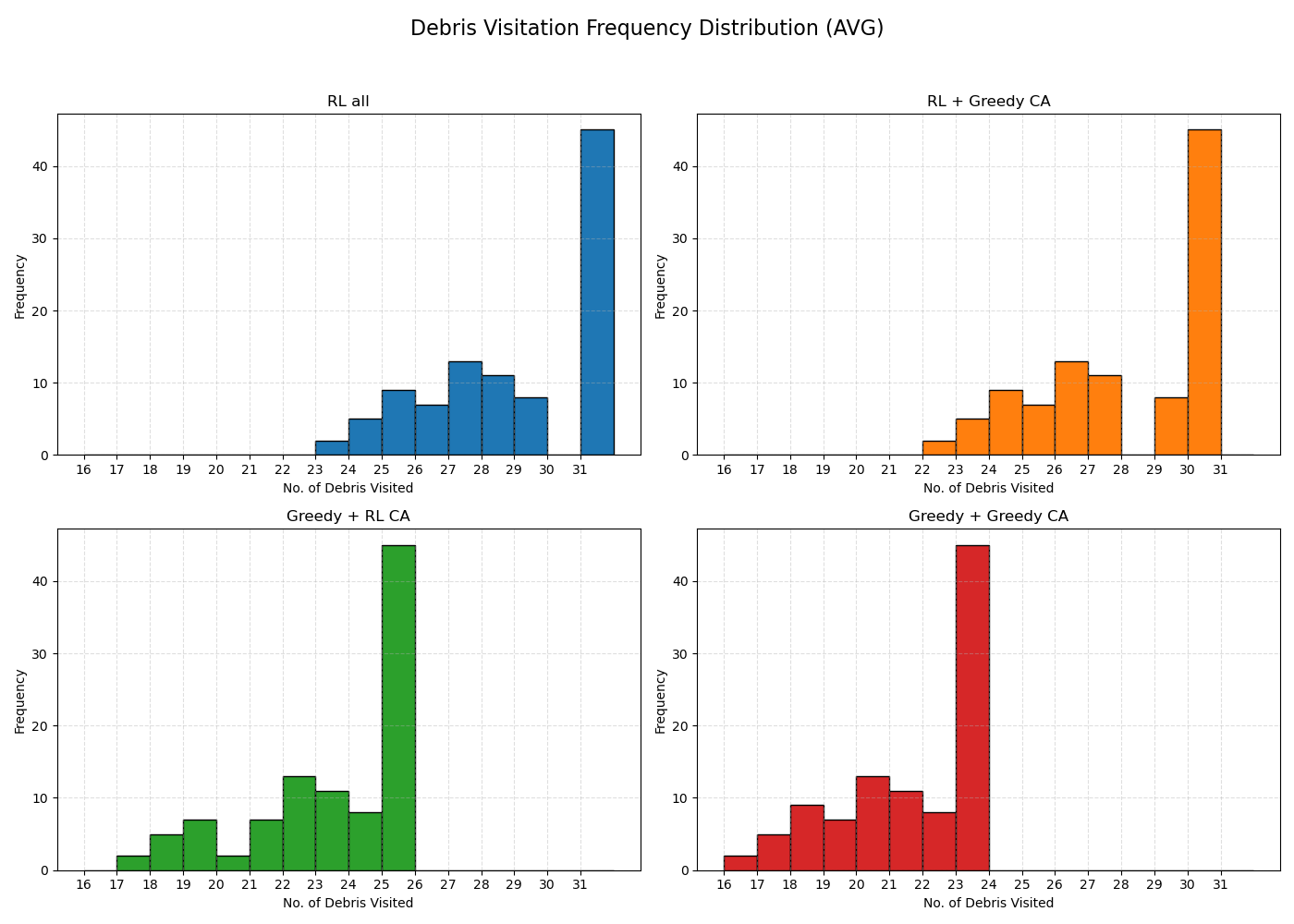}
    \caption{Debris Visited per Strategy (AVG) across 100 randomized test cases. RL-RL performs best across almost all scenarios.}
    \label{fig:avg_debris_plot}
\end{figure*}

A key performance metric is the number of debris objects successfully visited per mission. Figure~\ref{fig:avg_debris_plot} illustrates the average debris visited across 100 test cases, with ten trials each.

The RL-only strategy (blue) achieves higher debris coverage on average compared to hybrid methods, as observed across the evaluated test cases. Hybrid strategies, particularly those relying on greedy heuristics for visitation planning, tend to under perform due to their limited ability to anticipate future constraints and dynamically adapt to collision risks. Among the tested configurations, the Greedy-Greedy and Greedy-RL modes show lower average debris visitation, likely due to non-optimal initial sequencing and reactive collision avoidance, which together restrict mission efficiency even when intelligent maneuvering is partially employed.

\subsection{Evaluation Summary}

Table~\ref{tab:evaluation_summary} summarizes key performance statistics across all 10 iterations of an example test scenario.

\begin{table}[htbp]
\centering
\begin{tabular}{|l|c|c|c|}
\hline
\textbf{Evaluation Type} & \textbf{Average} & \textbf{Max} & \textbf{Min} \\
\hline
RL all              & 30.4 & 31 & 29 \\
RL + Greedy CA      & 29.5 & 31 & 28 \\
Greedy + RL CA      & 21.6 & 23 & 21 \\
Greedy + Greedy     & 19.3 & 23 & 17 \\
\hline
\end{tabular}
\caption{Evaluation summary for all 10 iterations of an 
example test case, showing the number of Debris being visited by each strategy.}
\label{tab:evaluation_summary}
\end{table}

The RL-RL configuration not only maximizes debris coverage but also ensures consistent performance across varying debris configurations and collision scenarios. The narrow performance gap between RL-RL and RL-Greedy CA validates the robustness of the learned visitation policy, even when paired with a simpler avoidance method. However, the reverse combination—Greedy + RL CA—fails to achieve comparable performance due to poor target prioritization.

\subsection{Policy Behavior: A Case Study}

To gain further insight into the agent’s learned behavior, we analyze two representative policy traces under different environmental conditions. In both scenarios, the agent successfully visited 31 debris targets within the allowed mission time and constraints. However, the decision-making patterns reflect meaningful adaptations based on collision risk:

\begin{quote}
\texttt{DEBRIS\_0 → Refuel → DEBRIS\_15 → CA\_BELOW\_DEBRIS\_37 → Refuel → DEBRIS\_5 → Refuel → ... → DEBRIS\_14}  
\hfill (Collision Probability = 33\%)
\end{quote}

\begin{quote}
\texttt{DEBRIS\_0 → Refuel → DEBRIS\_15 → DEBRIS\_21 → Refuel → ... → DEBRIS\_47}  
\hfill (Collision Probability = 0\%)
\end{quote}

These traces illustrate several key aspects of the policy's behavior:
\begin{itemize}
    \item \textbf{Robust planning under uncertainty}: When there was no collision, the agent had chosen a different route to complete its mission. However, even after facing an uncertain event like collision, it performed the avoidance maneuver and dynamically changed its route to get the same result as the other agent.
    \item \textbf{Efficient mission execution}: Despite the added complexity of CA and extra detours, the agent maintained high performance by completing its mission within limits.
\end{itemize}

This comparison highlights the policy’s ability to adaptively replan in response to dynamic hazards, while still optimizing for task completion. Such emergent behavior is not hard-coded but learned solely through interaction, reward shaping, and training exposure.

\subsection{Interpretation and Impact}

The high success rate of the RL-based policy underlines the importance of end-to-end learning in multi-objective space missions. Unlike greedy methods that optimize locally, the RL policy learns globally coherent strategies, balancing immediate rewards with long-term feasibility.

These findings have implications beyond ADR. The core principles—dynamic planning, collision adaptation, and refueling decisions—are transferable to satellite servicing, in-orbit inspections, or planetary lander missions.

Our results confirm the effectiveness of reinforcement learning for complex space mission planning. By jointly learning visitation sequences, refueling logic, and collision avoidance maneuvers, the agent achieves performance on par with or exceeding traditional baselines. This approach offers a scalable path forward for autonomous multi-debris rendezvous missions in increasingly congested orbital environments.

In future, we aim to extend this framework to:
\begin{itemize}
    \item More complex simulation scenarios.
    \item Multi-agent coordination between multiple debris-removal satellites.
    \item Online learning for in-situ policy updates.
    \item Integration with flight software for real-time deployment and space qualification.
\end{itemize}

\section{CONCLUSION}

This work presents a novel reinforcement learning-based mission planning framework designed for multi-debris rendezvous in congested orbital environments. By integrating dynamic collision avoidance, intelligent refueling logic, and adaptive sequencing through masked Proximal Policy Optimization, the agent consistently achieves near-optimal coverage in highly constrained scenarios.

Our approach demonstrates significant advantages over traditional greedy and hybrid baselines. The RL agent learns to anticipate collision risks, trigger safe detours, and manage fuel usage efficiently, resulting in robust, flexible mission behavior that generalizes well across randomized debris configurations.

The results validate reinforcement learning as a viable method for next-generation autonomous space operations. This framework is particularly relevant to future Active Debris Removal (ADR) initiatives, where safety, efficiency, and autonomy are critical.
The robustness and the adaptability of the RL agent would come in handy in various other space problems.

\section*{APPENDIX}
\addcontentsline{toc}{section}{APPENDIX}  

\subsection*{A. Mission Parameters}
\begin{itemize}
    \item \textbf{Initial Orbit and Refuelling Orbit:} 700 km circular orbit, 96° inclination.
    \item \textbf{Target Debris:} 50 objects sampled per episode randomly between 700 and 800 km.
    \item \textbf{Collision Zone Size:} $5 \times 5 \times 5$ km cuboid.
    \item \textbf{Collision Probability:} 33\% chance upon debris selection.
    \item \textbf{Clearance Requirement:} 5 km minimum for detour.
    \item \textbf{Refueling Logic:} Enabled after at least one debris rendezvous.
    \item \textbf{Fuel Capacity:} Normalized to 1.0 at start; reset to full on refuel calculated with reference to the maximum dV.
    \item \textbf{Max dV:} Set at 3 km/s.
    \item \textbf{Maximum Episode Duration:} Each test case has a duration of 7 days within which the agent has to visit as many debris as possible.
    \item \textbf{Episode Termination:} All debris visited, fuel exhausted, running out of time or invalid action.
    \item \textbf{Test Case Variation:} Each test case has generates 50 random debris. Each test case is tested for 10 iterations with each strategy as the collisions are completely randomized per iteration.
    \item \textbf{Assumptions:} We assume no effect of the J2 pertubations.

\end{itemize}

\subsection*{B. Reinforcement Learning Hyperparameters}
\begin{itemize}
    \item \textbf{Algorithm:} Proximal Policy Optimization (Masked PPO).
    \item \textbf{Learning Rate:} $5 \times 10^{-6}$.
    \item \textbf{Training Steps:} 10 million.
    \item \textbf{Batch Size:} 2048.
    \item \textbf{Discount Factor ($\gamma$):} 0.99.
    \item \textbf{Clipping Parameter ($\epsilon$):} 0.2.
    \item \textbf{Action Masking:} Enabled (invalid actions set to $-\infty$ logits).
    \item \textbf{Policy Network:} Multilayer Perceptron (MLP) with two hidden layers, 256 neurons each, \texttt{tanh} activation.
    \item \textbf{Framework:} Stable-Baselines3 + Poliastro.
    \item \textbf{Hyperparameter selection:} The hyperparameters were selected based on the optimum performance achieved in our previous works.
    
\end{itemize}

\subsection{C.Greedy heuristic parameters}
\begin{itemize}
    \item \textbf{Algorithm:} The greedy algorithm chooses the rendezvous with the lowest dv.
\end{itemize}

\subsection*{D. Plain MCTS Configuration}
\begin{itemize}
    \item \textbf{Algorithm:} Plain Monte Carlo Tree Search using UCT.
    \item \textbf{Exploration Constant:} $c_{\text{uct}} = 1.5$.
    \item \textbf{Simulations per Step:} 200.
    \item \textbf{Rollout Depth:} 15.
    
\end{itemize}

\subsection*{E. Environment Settings}
\begin{itemize}
    \item \textbf{Astrodynamics Libraries:} Poliastro, Astropy.
    \item \textbf{Collision Checking:} Discrete zone overlap at 60s intervals.
    \item \textbf{CA Maneuver Logic:} Elliptical detours—CA\_Above or CA\_Below.
    \item \textbf{Refuel Duration Penalty:} 2.5 hours added per refuel event.
    \item \textbf{Hardware:} MacBook Pro M1 Max, 64 GB RAM, Python 3.11.
\end{itemize}

\sloppy
\bibliographystyle{IEEEtran}
\bibliography{references}

\end{document}